%% file: main.tex
\documentclass[10pt,twocolumn,letterpaper]{article}

\usepackage{iccv}              
\usepackage{multicol}
\usepackage{multirow}
\usepackage{booktabs}
\usepackage{makecell}


\definecolor{iccvblue}{rgb}{0.21,0.49,0.74}
\usepackage[pagebackref,breaklinks,colorlinks,allcolors=iccvblue]{hyperref}

\def\paperID{2986} 
\def\confName{ICCV}
\def\confYear{2025}

\title{TRKT: Weakly Supervised Dynamic Scene Graph Generation with Temporal-enhanced Relation-aware Knowledge Transferring}

\author{Zhu Xu$^1$\quad Ting Lei$^1$\quad Zhimin Li$^2$\quad Guan Wang$^3$\quad Qingchao Chen$^4$\quad Yuxin Peng$^1$\quad Yang Liu$^{1}$\thanks{Corresponding author} \\
$^1$Wangxuan Institute of Computer Technology, Peking University\\$^2$ Tencent Inc.\quad $^3$ Baidu Inc. \quad $^4$ National Institute of Health Data Science, Peking University\\
{\tt\small xuzhu@stu.pku.edu.cn \{ting\_lei, qingchao.chen, pengyuxin, yangliu\}@pku.edu.cn}\\ \tt\small zhiminli.cn@outlook.com \tt\small wangguan15@baidu.com \\
}
\begin{document}
\maketitle
\input{sec/0_abstract_v2}    
\input{sec/1_intro_v3}

\input{sec/2_formatting}

\input{sec/3_finalcopy}
\input{sec/4_exp}
{
    \small
    \bibliographystyle{ieeenat_fullname}
    \bibliography{main}
}

\end{document}

%% file: sec/0_abstract_v2.tex
\begin{abstract}
Dynamic Scene Graph Generation (DSGG) aims to create a scene graph for each video frame by detecting objects and predicting their relationships. Weakly Supervised DSGG (WS-DSGG) reduces annotation workload by using an unlocalized scene graph from a single frame per video for training.
Existing WS-DSGG methods depend on an off-the-shelf external object detector to generate pseudo labels for subsequent DSGG training. However, detectors trained on static, object-centric images struggle in dynamic, relation-aware scenarios required for DSGG, leading to inaccurate localization and low-confidence proposals.
To address the challenges posed by external object detectors in WS-DSGG, we propose a Temporal-enhanced Relation-aware Knowledge Transferring (TRKT) method, which leverages knowledge to enhance detection in relation-aware dynamic scenarios. 
TRKT is built on two key components:(1)\textit{Relation-aware knowledge mining}: we first employ object and relation class decoders that generate category-specific attention maps to highlight both object regions and interactive areas. Then we propose an Inter-frame Attention Augmentation strategy that exploits optical flow for neighboring frames to enhance the attention maps, making them motion-aware and robust to motion blur. This step yields relation- and motion-aware knowledge mining for WS-DSGG. (2) we introduce a Dual-stream Fusion Module that integrates category-specific attention maps into external detections to refine object localization and boost confidence scores for object proposals. 
Extensive experiments demonstrate that TRKT achieves state-of-the-art performance on Action Genome dataset. Our code is avaliable at \url{https://github.com/XZPKU/TRKT.git}.

\end{abstract}

%% file: sec/1_intro_v3.tex
\section{Introduction}
\label{sec:intro}

Dynamic Scene Graph Generation (DSGG) aims to represent complex visual scenes in a video sequence as structured graphs, with nodes representing object instances while edges capture the relationships between objects, which is valuable for visual-language tasks like Human Object Interaction Detection~\cite{lei2024exploring, Lei_2023_ICCV, Lei_2024_CVPR}, Visual Grounding~\cite{10.1145/3664647.3681660, zheng2024TFTG}, Visual Question Answering~\cite{3DVQA_2024_AAAI,3DSyn}. However, annotating for video scene graphs is highly challenging and resource-intensive, hindering the scaling of DSGG. To address it, weakly supervised DSGG~\cite{chen2023video} (WS-DSGG) has been proposed, which relies on only one unlocalized scene graph from one frame of video as supervision, making it a practical solution for expanding DSGG to more complex video data.

\input{figures/fig3_teaser2}
\input{figures/fig1_teaser}

Existing WS-DSGG approaches, such as the state-of-the-art model PLA~\cite{chen2023video}, rely on off-the-shelf external object detectors to generate object labels, which are then used to construct pseudo-localized scene graphs for training a fully supervised DSGG model. However, these external detectors often struggle with the DSGG task due to two key limitations: (1) video frames contain dynamic motion and potential blurring, whereas external detectors are trained on static, object-centric images, making them ill-suited for such scenarios; (2)
external detectors are trained solely on object annotations, overlooking crucial relational cues necessary for dynamic scene graph data, leading to biased detection bounding boxes that miss interactive object boundary areas. These limitations lead to inaccurate object localization and missing detections, ultimately degrading the quality of pseudo labels and hindering final DSGG performance.
To systematically assess the impact of object detection and predicate prediction on DSGG performance, we conduct an evaluation using PLA. As shown in Fig.~\ref{fig:teaser2}, combining PLA’s object detection results with oracle relation labels leads to a modest 0.9\% improvement in DSGG performance. In stark contrast, substituting PLA’s object detections with oracle detections—thus providing accurate object pairs for training—yields a substantial 61.9\% performance boost. These findings highlight the critical role of external detector quality in WS-DSGG performance.

To address the challenges mentioned above, we propose an approach termed Temporal-enhanced Relation-aware Knowledge Transferring (TRKT) for the WS-DSGG task. Our method strategically targets the identified external detection weaknesses by mining and integrating knowledge, with the aim to make the detection results both relation-aware and motion-aware. Specifically, we employ a class-sensitive object and relation decoder the extract class-sensitive attention maps for potential objects, using relation-aware weak supervision tailored to the specific context of the WS-DSGG task. We introduce both object and relation tokens in the decoder to explicitly encapsulate objects as well as the relations between objects within the scene, enhancing the model's ability to capture and prioritize the relation regions of corresponding objects, thus improving the overall quality of the scene graph.
Furthermore, we utilize optical flow information to provide a temporal cue on object movement across video frames, which is crucial for maintaining coherence in object tracking. We then augment the class-sensitive attention maps between adjacent frames with these temporal cues, alleviating the problems of blurring and occlusion that arise in dynamic scenes.

To this end, we generate class-sensitive attention maps that are both relation-aware and motion-aware. However, effectively transferring the knowledge embedded in these attention maps to enhance external detection results presents a significant challenge. To address this, we propose a Dual-stream Fusion Module (DFM), which consists of two key components:
(1) Confidence Boosting Module (CBM): This module mitigates the issue of low-confidence detections by leveraging class-sensitive attention maps to reassess and refine the confidence scores of external detections. For each high-confidence category identified by the class decoder, CBM validates the corresponding detection scores, reducing the risk of missed detections caused by low-confidence predictions. As illustrated in~\cref{fig:teaser}, CBM enhances the confidence score of the detected ``laptop” in the right image.
(2) Localization Refinement Module (LRM): This module integrates temporal and relational information from class-sensitive attention maps to refine bounding box coordinates, addressing localization inaccuracies. This process employs Weighted Box Fusion~\cite{solovyev2021weighted} to enhance object localization. As shown in~\cref{fig:teaser}, LRM corrects an undersized ``person" bounding box by incorporating semantic attention cues.

In summary, the main contributions are as follows:
(1) We demonstrate the significant impact of object detection quality on the performance of WS-DSGG, highlighting the key limitations of existing methods.
(2) We propose the novel approach Temporal-enhanced Relation-aware Knowledge Transferring (TRKT), which utilizes class-sensitive knowledge that are both relation-aware and motion-aware, to enhance the object detection specifically for WS-DSGG.
(3) We further introduce the Dual-stream Fusion Module (DFM) to incorporate the above attention maps to improve the external detection results, which comprises Localization Refinement Module (LRM) to improve bounding box accuracy and the Confidence Boosting Module (CBM) to dynamically adjust confidence scores.
(4) We validate the effectiveness of our TRKT framework through extensive experiments, underscoring the critical role of improved object detection accuracy in enhancing WS-DSGG performance.

%% file: figures/fig3_teaser2.tex
\begin{figure}[t]
    \includegraphics[width=\linewidth,height=6cm]{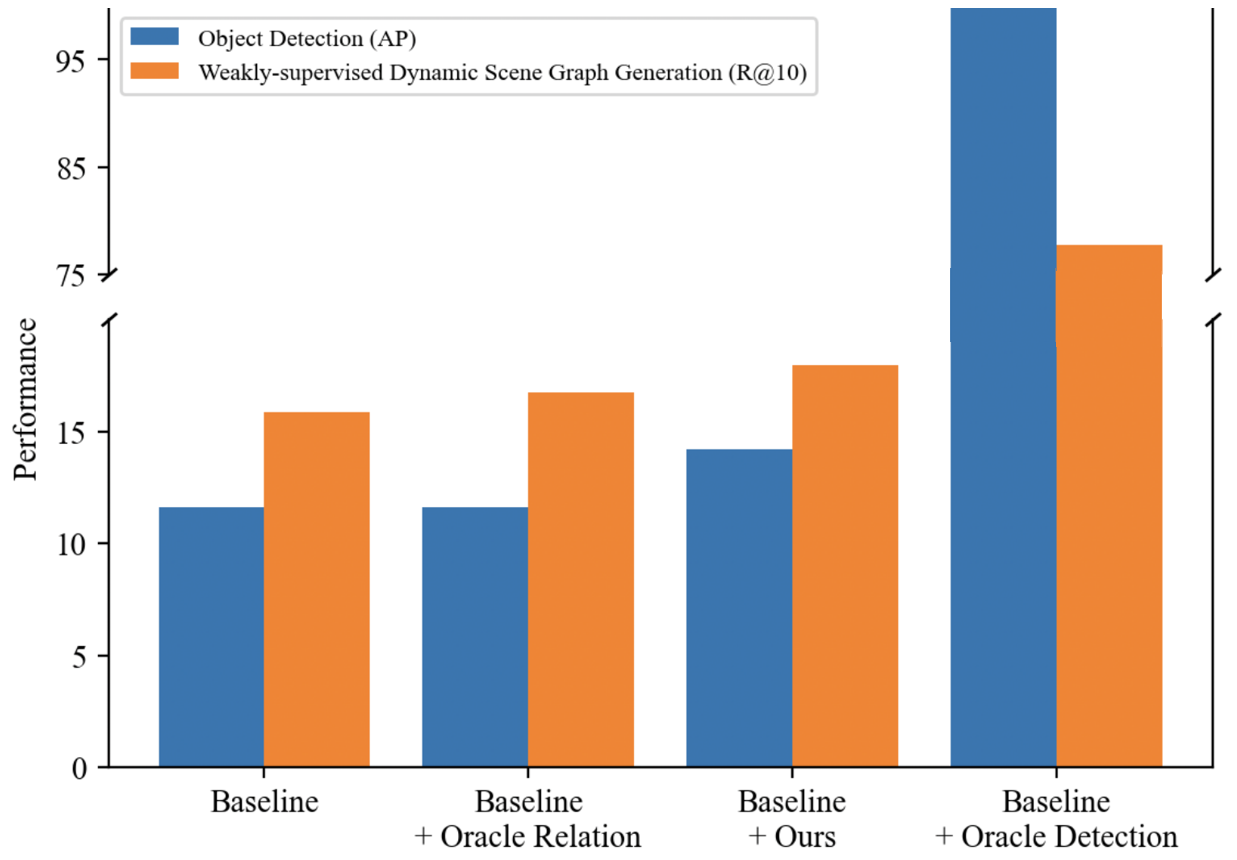}
    \caption{Performance Comparison between different methods in object detection and WS-DSGG.}
\label{fig:teaser2}
\end{figure}

%% file: figures/fig1_teaser.tex
\begin{figure}[t]
    \includegraphics[width=\linewidth]{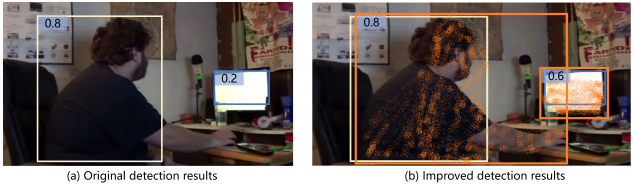}
    \caption{Comparison between (a) existing off-the-shelf object detection results and (b) improved object detection results with visual cues under weak supervision.}
    \vspace{-3mm}
\label{fig:teaser}
\end{figure}

%% file: sec/2_formatting.tex
\section{Related Work}
\subsection{Dynamic Scene Graph Generation}
\vspace{-1mm}
 DSGG is a spatial-temporal scene understanding task that aims to leverage temporal context to analyze the objects along with their dynamic relationships~\cite{ji2020action}. Two distinct task formulations exist for DSGG based on their output format:
(1) Traclet-level~\cite{Gao_2021,gao2022classificationthengroundingreformulatingvideoscene,gao2023compositionalprompttuningmotion}: each node of output scene graph represents one object traclet in the input video. 
(2) Frame-level~\cite{9878927,feng2023exploiting,cong2021spatial,teng2021target,arnab2021unified}: Each node of scene graph represents one object bounding box, and each video frame should predict its scene graph respectively.
We adopt the frame-level formulation as it provides finer-grained scene graph representations, allowing more precise modeling of dynamic relations at each time step and better handling rapidly changing relations in complex scenes.
Despite the advancements~\cite{cong2021spatial, feng2023exploiting,teng2021target, arnab2021unified} in frame-level DSGG, such full-supervised training manner relies heavily on extensive manual annotations, including object category, bounding boxes and relation category, which is very costly.  Therefore, we investigate on the Weakly-supervised DSGG, which only demands one single-frame unlocalized scene graph for each video, to effectively alleviate such heavy reliance.

\subsection{Weakly Supervised DSGG}
\vspace{-1mm}
Weakly Supervised DSGG(WS-DSGG) is proposed to alleviate the labor-intensive full scene graph annotation, which only utilizes single-framed unlocalized scene graph as supervision for each input video. 
PLA\cite{chen2023video} first formulates this task and proposes to utilizes an external detector to detect bounding boxes for objects, which are then assigned to image-level class annotations to create pseudo scene graphs for each frame.
Then it train a fully-supervised Image Scene Graph Generation (ISGG) model with these pseudo scene graphs, which are subsequently employed to train DSGG model in full-supervised manner. Building upon this pipeline, NL-VSGG\cite{kim2025weaklysupervisedvideoscene} further proposes to use video captions as weaker supervision to construct pseudo scene graphs.
However, these previous works all rely on external object detectors in the early stage, which hampers the ultimate DSGG performance. To mitigate this, we propose incorporating object and relation knowledge, along with temporal information, to boost the detection accuracy early on. These strategies operate synergistically to enhance object detections, thereby improving DSGG performance.
\subsection{Weakly Supervised Object Detection}
\vspace{-1mm}
Weakly Supervised Object Detection (WS-OD) aims to train an object detector with only object class labels. Most advancements\cite{xu2022h2fa,9009008,yang2019preciseendtoendweaklysupervised,yang2019activitydrivenweaklysupervised, E210455, 10410600, 9996117, Yang_2024_CVPR} for WS-OD adopt a multi-stage pipeline. $\rm H^2FA$-RCNN\cite{xu2022h2fa} first generates object proposals under the supervision of image labels, then employs multi-level feature alignment to filter proposals. C-MIDN\cite{yang2019preciseendtoendweaklysupervised} proposes a coupling strategy to mine more complete proposals. Notably, different from WS-DSGG, WS-OD is not allowed to use external object detectors to acquire external detection results, so our design, which targets refining external detection results by integrating relation-aware knowledge, is not applicable to WS-OD task.

%% file: sec/3_finalcopy.tex
\section{Method}
\input{figures/fig2_pipeline}
\subsection{Problem Formulation}
Weakly supervised DSGG is to detect all visual relations within a video sequence $V=\{I_{1}, I_{2}, \cdots, I_{N_v}\}$, where $N_v$ is the frame number of video. The only annotation $\Tilde{G}_T$ is an unlocalized scene graph from one frame of video, where $\Tilde{G}_T=\{(s_i, o_i, p_i)_{i=1}^{G}\}$, $T$ denotes the index of the annotated frame within the video sequence $V$, $s_i$, $o_i$, and $p_i$ representing the category labels for $i^{th}$ subject, object, and predicate, and G is the triplet number in the annotation. The target is to generate localized scene graph predictions for each video frame $I_t$.
\subsection{Baseline Revisited}
\vspace{-1mm}
\label{subsec:pla_baseline}
We adopt PLA\cite{chen2023video} for WS-DSGG task as our baseline.
We adopt the most advanced approach targeting for WS-DSGG task, PLA~\cite{chen2023video}, as our baseline. Considering that most of the frames within the video lack training supervision, PLA proposes a Pseudo Label Assignment-based approach, then utilizes these pseudo labels to train one DSGG model in a fully supervised manner. To generate object pseudo labels, it adopts an external object detector to acquire object proposals for all frames, denoted as $D_e$. Specifically, for the frame with unlocalized scene graph annotation, PLA matches the detected object proposals with the scene graph according to their categories, acquiring pseudo-localized scene graph $G_{pseudo}$. Further, PLA introduces a strategy based on object category information to assign $G_{pseudo}$ to other frames and construct pseudo scene graph labels for full-supervised DSGG model training.

\subsection{Error Analysis For External Detection}
\vspace{-1mm}
Despite significant progress with PLA, we empirically observe that external object detection still falls short in meeting the specific requirements of WS-DSGG, and the overall model performance is heavily dependent on the quality of object detection results (see Fig.~\ref{fig:teaser2}). To address this, we first conduct an error analysis on the external detection results, following the TIDE criteria~\cite{bolya2020tidegeneraltoolboxidentifying} to categorize errors into six main types: classification error, localization error, both classification and localization error, duplicate error, background error\footnote{Background error denotes detecting background as foreground object, which is caused by inaccurate bounding box localization.}
, and missed ground truth (GT) error. 
Our empirical analysis reveals that localization errors and background errors, both of which stem from inaccurate object localization, together account for 39.83\% of total errors. Missed GT errors, resulting from low-confidence detection proposals, account for 59.43\%. Based on these findings, we identify that inaccurate object localization and low-confidence detections are the primary issues in external detection. Our method focuses on addressing these two problems by guiding external detection to be both relation-aware and motion-aware, ultimately improving the DSGG task performance.

\subsection{Overview}

The overall pipeline is depicted in \cref{fig:pipeline} and comprises two essential components.
The first component focuses on mining and transferring temporal-enhanced relation-aware knowledge tailored for WS-DSGG. We utilize an image encoder to process the input image into patches, which are then analyzed by separate object and relation class decoders to produce attention maps that emphasize relevant areas for objects and their interactive relations. This encoder and decoder are supervised by only image-level class labels. These class-sensitive attention maps contain object semantics and potential relational contexts, thus enhancing the model's ability to recognize and understand complex relations within the data. The cross-framed optical flow is further utilized to provide temporal cues to augment the attention maps by alleviating blur and occlusion problems. To this end, we yield class-specific attention maps that are both relation-aware and motion-aware, which are thus utilized to complement external detectors for WS-DSGG.

Further, to maximize the utility of the class-specific attention maps for improving external detection results, we introduce a dual-stream fusion module featuring Localization Refinement Module (LRM) and the Confidence Boosting Module (CBM). The LRM refines bounding box coordinates of external detection by employing attention maps to pinpoint critical object regions, while CBM amplifies the confidence scores of object categories identified by the class decoder. This dual approach effectively mitigates the influence of biases that existed in external detection results, resulting in more reliable object detection outcomes.

The detection outputs are then organized as pseudo-scene graphs following  PLA~\cite{chen2023video} as described in the Baseline section. Ultimately, the trained DSGG model generates a scene graph for each frame of the input video. In the following, we further explain our \textbf{Relation-aware Knowledge Mining} and \textbf{Dual-stream Fusion Module}.

\subsection{Relation-aware Knowledge Mining}
\vspace{-1mm}
The external detectors trained on static and object-annotation-only data exhibit sub-optimal detection performance on DSGG data that requires relation understanding and encompasses dynamic motion. So to address these issues, we propose to mine relation-aware knowledge tailored for the DSGG task. Specifically, we use unlocalized annotations to train object and relation class decoders, generating class-sensitive attention maps that are aware of both object and their relation regions. Further, we use neighboring frames and optical flow information to create pseudo-attention maps to reduce blur and occlusion, enhancing the attention maps to be motion-aware.

\noindent\textbf{Class-Sensitive Attention Map Generation.}
Since we only have image-level object and relation labels, our goal is to derive visual cues specific to different objects and their relationships from these annotations. To achieve this, we propose a method that leverages a transformer-based encoder-decoder architecture. This architecture uses object and relation tokens to focus attention on the corresponding object and relation regions in the attention maps, revealing positional cues for each instance.
We first encode input image $I_{t}$ into image patch features $x_{patches} \in \mathbb{R}^{N \times D} $, where $N$ is the patch number and $D$ is the feature dimension. Then, to attend to specific regions highly related to each object category, we equip the object class encoder with object tokens $x_{obj} \in \mathbb{R}^{C_{obj} \times D}$, where $C_{obj}$ is the object class number. Further, considering that some object categories exhibit significant interactive behaviors, we provide relation  tokens $x_{rel} \in \mathbb{R}^{C_{rel} \times D}$ in relation class decoder to attend to regions encompassing relationship activities, where $C_{rel}$ is the relation class number. Then for attention layers within each class decoder, we formulate the attention calculation as follows,
\begin{equation}
\label{eq:rel_attn_ext}
\begin{aligned}
    &x_{tgt}=\text{CA}( x_{tgt}, x_{patches}) \\
    &=\text{Softmax}\left(  (x_{tgt} W_q) (z W_k)^T) / \sqrt{D}  \right) z W_v
    =\mathbf{A}_{tgt}(z W_v) \\
\end{aligned}
\end{equation}
where $tgt$ could be $obj$ and $rel$, $\text{CA}$ denotes cross-attention layer,  $z = [ x_{tgt}, x_{patches}]$  
indicates the concatenated embeddings, and $W_q, W_k, W_v$ are the query, key, and value projection layers, respectively. The rationale behind concatenating $x_{tgt}$ and $x_{patches}$ for attention calculation is to supply comprehensive contextual cues from both image features and object or relation tokens for accurately attending region of interest.
$\mathbf{A}_{tgt} \in \mathbb{R}^{C_{tgt} \times (C_{tgt}+N)}$ represents the attention matrix. As the tokens $x_{tgt}$ are adopted to capture class-specific visual clues, we derive $\mathbf{A}_{tgt}^* \in \mathbb{R}^{C_{tgt} \times h \times w}$ from $\mathbf{A}_{tgt}$ by slicing and reshaping, where $N = h*w$, representing the attention between object/relation tokens and image patch features. 

To guide the decoders construct more accurate attention maps $\mathbf{A}^{*}_{obj}$ and $\mathbf{A}^{*}_{rel}$, we utilize the unlocalized scene graph $\Tilde{G}_T=\{(s_i, o_i, p_i)_{i=1}^{G}\}$ to form object and relation image labels as supervision. Specifically, we aggregate all the object categories $\{s_{i}\}_{i=1}^{G}$ and $\{o_{i}\}_{i=1}^{G}$ within $\Tilde{G}_{T}$ to construct the multi-class image object label vector $y = [y_1, y_2, ..., y_{C_{obj}} ]^T $. Similarly, we construct multi-class image relation label vector $p = [p_1,p_2,..., p_{C_{rel}}]$ by aggregating $\{p_{i}\}_{i=1}^{G}$ in $\Tilde{G}_T$.  Then for decoded object and relation visual feature $ x_{obj}, x_{rel}$, extra linear layer $W_o, W_r \in \mathbb{R}^{D\times 1}$ are introduced to project corresponding tokens into classification logits $s_o \in \mathbb{R}^{C_{obj}\times 1}$ and $s_r \in \mathbb{R}^{C_{rel}\times 1}$, where $C_{obj}$ and $C_{rel}$ indicating the number of image object labels and relation labels.
The loss for classification is formulated as 
\begin{equation}
\begin{aligned}
    \mathcal{L}&=\mathcal{L}_{obj} + \mathcal{L}_{rel} =  \mathcal{L}_{\operatorname{BCE}}(s_o,y) + \mathcal{L}_{\operatorname{BCE}}(s_r,p)\\
    &= \mathcal{L}_{\operatorname{BCE}}(x_{obj} W_o,y) + \mathcal{L}_{\operatorname{BCE}}(x_{rel} W_r,p)
\end{aligned}
\end{equation}
where $\mathcal{L}_{\operatorname{BCE}}(\cdot)$ indicates the binary cross-entropy loss.

\noindent\textbf{Class-sensitive Attention Fusion.}
The object and relation tokens capture relation-aware knowledge about the spatial locations of object categories from different perspectives. The object tokens focus on class-specific features, highlighting regions representing each category, while the relation tokens emphasize interactive behaviors, covering regions where objects may interact. 
To generate more accurate class-sensitive attention maps, we propose to fuse relation attention maps $A_{rel}^{*} \in \mathbb{R}^{C_{rel} \times h \times w}$ into object attention maps $A_{obj}^{*} \in \mathbb{R}^{C_{obj} \times h \times w}$, since class-sensitive attention maps ultimately target for object localization, focusing specifically on each object instance while being sensitive to the interactive boundaries for each object.
Formally, we first calculate the similarity $S$ between $\mathbf{A}_{rel}^{*}$ into $\mathbf{A}_{obj}^{*}$. A higher similarity between object and relation attention map indicates that the corresponding object and relation are not only spatially close, but also share a similar pattern, thus the object probably involves the interaction. So we fuse $\mathbf{A}_{rel}$ into $\mathbf{A}_{obj}$ according to $S$, formulated as
\begin{equation}
\begin{aligned}
    S = \mathbf{A}_{obj}^{*} \cdot (\mathbf{A}_{rel}^{*})^{T}
     \quad \mathbf{A}_{obj}^{*'} = \text{norm}(\mathbf{A}_{obj}^{*} + S \cdot \mathbf{A}_{rel}^{*})
\end{aligned}
\end{equation}
where $S \in \mathbf{R}^{C_{obj} \times C_{rel}}$, $\mathbf{A}_{obj}^{*'} \in \mathbf{R}^{C_{obj} \times h\times w}$, and $\text{norm}$ is normalization operation. 

\noindent\textbf{Inter-frame Attention Augmentation}
To further handle the potential challenges of motion blur and occlusion in video sequences and enable attention maps motion-aware, we propose a strategy Inter-frame Attention Augmentation (IAA), which adopts cross-framed optical flow information as temporal cues. For each frame $I_i$ (i=2,3,...$N_v$) within video sequence $V$, we adopt the neighboring frame $I_{i-1}$ to provide extra information, rescuing for possible object mis-detections caused by blur and occlusion in $I_i$. Specifically, we employ \cite{teed2020raftrecurrentallpairsfield} to obtain the inter-frame optical flow $OF_{i-1,i}$, and acquire the class-sensitive attention maps $\mathbf{A}_{obj_{i-1}}^{*'}$ for $I_{i-1}$ with the same relation-aware knowledge mining process. Then we warp $\mathbf{A}_{obj_{i-1}}^{*'}$ according to flow field $OF_{i-1,i}$ to generate pseudo-attention maps $\mathbf{PA}_{obj}^{*'}$ for $i^{th}$ frame $I_i$, which enriched of temporal cues about the movement of objects.

By acquiring class-sensitive attention maps $\mathbf{A}_{obj}^{*'}$ and $\mathbf{PA}_{obj}^{*'}$ that are relation- and motion-aware, highlight significant regions for each object along with its potential interaction semantic. We then detailedly introduce how to use $\mathbf{A}_{obj}^{*'}$ and $\mathbf{PA}_{obj}^{*'}$ to improve external detection results quality in following Dual-stream Fusion Module subsection.


\subsection{Dual-stream Fusion Module}
\label{3_6}
\vspace{-1mm}
To alleviate the influence of low quality external detection, we propose to incorporate relation-aware class-sensitive knowledge, which is formulated as class-sensitive attention maps $\mathbf{A}_{obj}^{*'}$ and temporal-enhanced version $\mathbf{PA}_{obj}^{*'}$, through Dual-stream Fusion Module (DFM). DFM comprises of Localization Refinement Module (LRM) to improve the localization quality for external object proposals, and Confidence Boosting Module (CBM) to enhance the confidence score for objects deemed to exist within the image, thus remedying possible missing object detection.

\noindent\textbf{Localization Refinement Module.}
 \input{figures/fig6_lrm}
 The process of Localization Refinement is shown in Fig.~\ref{fig:LRM}, the external detection results and object proposals from class-sensitive attention maps are utilized to acquire refined detection results (green one in the right bottom of Fig.~\ref{fig:LRM}). To refine external detection results $D_e$, we inject relation-aware class-sensitive knowledge within $\mathbf{A}_{obj}^{*'}$. Firstly, we obtain internal object proposals $D_a =f(\mathbf{A}_{obj}^{*'})=\{(b_i, c_i, s_i\}_{i=1} ^ n\} \in \mathbb{R}^{n\times 6}$ based on a threshold-based algorithm $f(\cdot)$, where $b_i \in \mathbf{R}^{4}$ is detected bounding box, $s_i \in \mathbf{R}^{1}$ is the confidence score calculated as the mean attention score within corresponding attention map inside $b_i$, $c_i \in \mathbf{R}^{1}$ is the category of object, and $n$ represents the number of detected objects.
 
Then, we integrate $D_a$ with $D_e$ to acquire more accurate object bounding boxes with an object proposal fusion algorithm $F(\cdot)$, which is defaulted as Weighted Box Fusion~(WBF)~\cite{solovyev2021weighted}.  Specifically, WBF ranks detection results within same category by confidence, and builds clusters according to IoU. The final results are obtained from generated clusters, while the bounding box and scores are acquired from weighted sum operation. The fusion process is represented as follows:
\begin{equation}
\label{eq4}
\begin{aligned}
    D_1 = \mathcal{F}(D_a, D_e) 
        = \mathcal{F}(f(\mathbf{A}_{obj}^{*'}), D_e)
\end{aligned}
\end{equation}

\noindent\textbf{Confidence Boosting Module.}
Though the object localization is consequently improved by LRM, the problem of possible low confidence within some bounding boxes still hinders the object detection quality, leading to object missing when the confidence score is lower than a certain threshold. This originates from the domain shift from the pre-trained object detection dataset to the WS-DSGG dataset. Considering that ${\mathbf{A}_{obj}}^{*'}$ are enriched with relation-aware category-sensitive knowledge as they are generated under fully supervised training of image category labels, we complement external detection results by highlighting salient regions with high scores of class-specific attention maps and integrating them within external detection results, yielding high confidence for the corresponding proposals and remedying for possible object missing. We propose Confidence Boosting Module~(CBM) to implement it. As shown in Fig.~\ref{fig:CBM}, the scores for external detection results are boosted by our attention maps. Since CBM aims to enhance the confidence score for objects that ought to exist within the image, we only select object categories with a high probability during implementation, and the object classification logits $s_o$ serve as criteria. Formally, for each object category $c_i$ and its corresponding attention map ${\mathbf{A}_{obj}}^{*i'}$, if its logits $s_o^{i}$ higher than threshold $\phi$, we select the object $\{b,c,s\}$ from external detector, which owns the highest confidence among all object proposals of $c_i$ category. Then we construct external attention $A_{ext}$, where the value inside bounding box $b$ is $s$ and otherwise is $0$. Further, we add $A_{ext}$ with ${\mathbf{A}_{obj}}^{*i'}$, followed by a normalization operation, yielding augmented attention map for category $c_i$. 
Enhanced attention maps subsequently facilitate  object detection results $D_2$ via the threshold algorithm: 
\begin{equation}
\label{eq5}
    D_2 = f(\text{CBM}(\mathbf{A}_{obj}^{*'}, D_e))
\end{equation}
Then we fuse the object detection outcomes $D_1$ and $D_2$ to obtain refined object detection results $D = \mathcal{F}(D_1, D_2) $, which simultaneously harvest the accuracy improvement as well as confidence score boosting. Furthermore, to endow the detection results of temporal cues and alleviate potential blurring and occlusions in frames, we duplicate the operation of Eq.~\ref{eq4} and Eq.~\ref{eq5} with pseudo-attention maps $\mathbf{PA}_{obj}^{*'}$ upon $D$ to obtain ultimate detection results $D'$, which is adopted as upgraded object detection results to acquire pseudo scene graph labels as introduced in Baseline. We empirically find that injecting knowledge from relation-specific towards motion-specific, i.e., adopting $A_{obj}^{*'}$ and $PA_{obj}^{*'}$ in a sequential manner, benefits external detection quality the most. With these pseudo labels owing better object detection quality, the DSGG model performance is thoroughly improved. 

\input{figures/fig5_cbm}


%% file: figures/fig2_pipeline.tex
\begin{figure*}[t]
    \centering
    \includegraphics[width=0.8\linewidth]{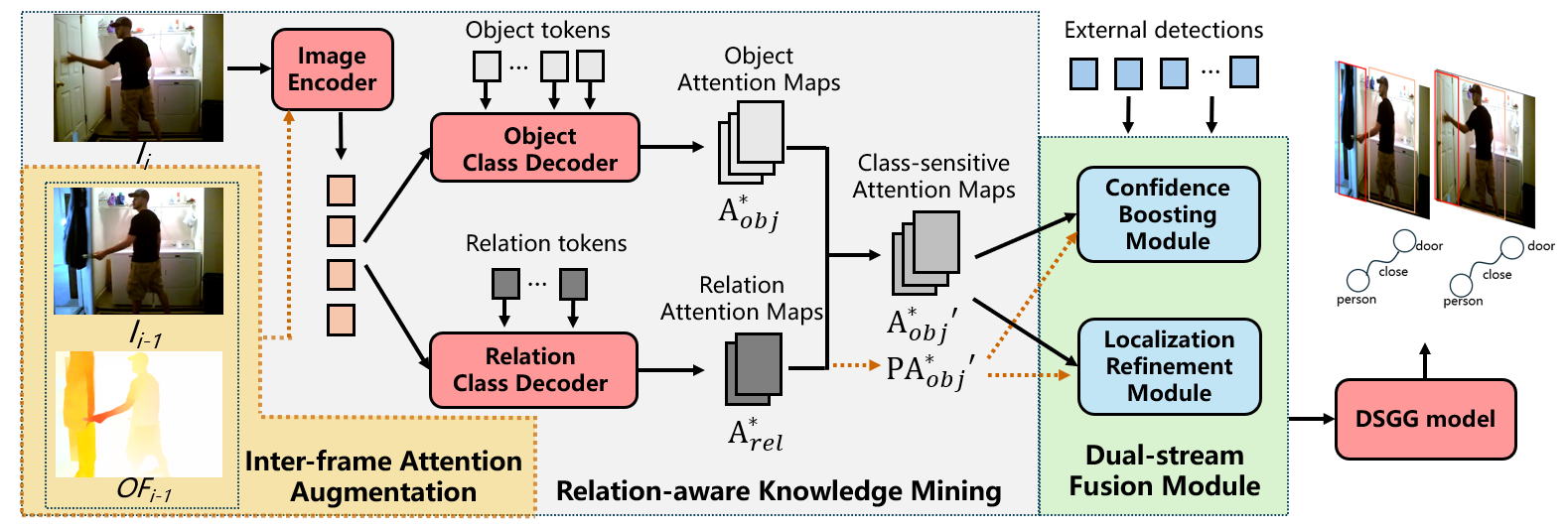}
    \vspace{-3mm}
    \caption{\textbf{Overall framework of our method TRKT}. 
    TRKT comprises two integral phases. During the \textbf{Relation-aware Knowledge Mining} phase, \textit{Object and Relation Class Decoder} separately generate attention maps that focus on specific object and relation semantic regions, and then fuse together to construct class-sensitive attention maps. Further, \textit{Inter-frame Attention Augmentation (IAA)} adopts previous frame equipped with cross-framed optical flow to generate pseudo attention maps aware of motion. Then \textbf{Dual-stream Fusion Module} uses class-sensitive attention maps to refine external detection results. \textit{Localization Refinement Module (LRM)} improves bounding box accuracy, while the \textit{Confidence Boosting Module (CBM)} boosts the confidence score for object proposals through attention projection.  Refined detection results are utilized to generate a pseudo scene graph for DSGG model training. }
\label{fig:pipeline}
\end{figure*}

%% file: figures/fig6_lrm.tex
\begin{figure}[!t]
    \centering
    \includegraphics[width=0.9\linewidth]{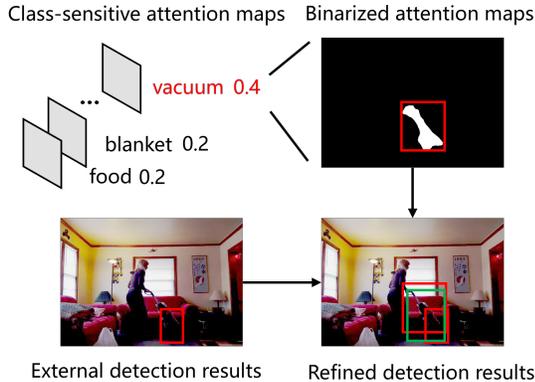}
    \caption{\textbf{Visualizations of Localization Refinement Module.}}
    \vspace{-5mm}
\label{fig:LRM}
\end{figure}

%% file: figures/fig5_cbm.tex
\begin{figure}[!t]
    \centering
    \includegraphics[width=0.9\linewidth]{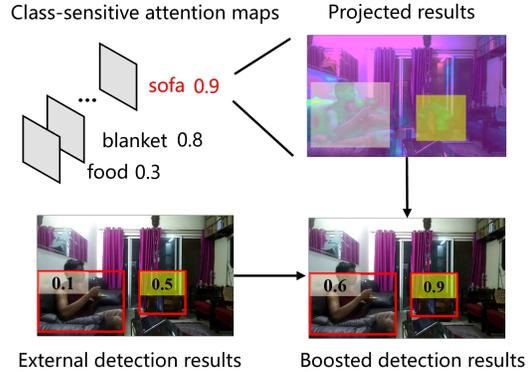}
    \caption{\textbf{Visualizations of Confidence Boosting Module.}}
    \vspace{-5mm}
\label{fig:CBM}
\end{figure}

%% file: sec/4_exp.tex
\input{tables/2_sota_det}
\input{tables/1_sota_v2}
\section{Experiments}

\label{sec:exp}

\input{tables/5_abla_source}

\input{tables/6_abla_fusion}

\input{tables/7_abla_token}

\subsection{Experimental Setting}
\noindent\textbf{Dataset and Evaluation Metrics.}
For dataset, we follow PLA~\cite{chen2023video} to evaluate our method on the Action Genome (AG) dataset~\cite{ji2020action}, which annotates $234, 253$ frame scene graphs for sampled frames from around 10K videos. The annotations encompass $35$ object categories and $25$ predicates. We also conduct experiments on VidVRD\cite{shang2017video} to verify our generalization towards different WS-DSGG task setting. For evaluation metrics, we evaluate our method in scene graph detection (SGDET) task in terms of the limitation that only unlocalized scene graph annotation is available in WS-DSGG.
SGDET aims to detect the relational object pairs and predict the relations between them.
All tasks are evaluated using the widely adopted Recall@K metrics (K = [10, 20, 50]), following the \textit{With Constraint} and \textit{No Constraint} settings~\cite{wang2022dynamic,ji2021detecting,li2022dynamic}.
Average Precision and Average Recall are also chosen as evaluation metrics for object detection performance, which is to gauge the effectiveness of the improvement for external object detection results.

\noindent\textbf{Implementation Details.}
For the external detector, we follow baseline PLA~\cite{chen2023video} to adopt VinVL~\cite{zhang2021vinvl} with backbone ResNeXt-152 C4. 
Our model is optimized by AdamW optimizer with a batch size of 4 on two GeForce RTX 3090 GPUs. We train our encoder and decoder for 20 epochs.

\subsection{Comparison with State of the Arts}
We first compare object detection results with our baseline PLA's original detection results, which are shown in Tab.~\ref{tab:sota_det}. Our proposed method averages an improvement of 13.0\%/1.3\%
for Average Precision(AP) and Average Recall(AR)\footnote{we report the averaged AP/AR for maxDets=1/10 by default, subsequent as well unless otherwise specified.} 
, validating that our approach effectively enhances object detection performance. 

Then for WS-DSGG, we compare with the only current work PLA~\cite{chen2023video} in  Tab.~\ref{tab:sota}.  PLA generates pseudo scene graph labels and then trains the DSGG model in fully-supervised manner.
As shown, we yield performance gains across all metrics compared to PLA (1.72\%/2.42\%
under \textit{With/No Constraint} scenario), which indicates that with improved object detection results, the generated pseudo scene graph labels quality are therefore improved, thus harvesting performance gain in ultimate DSGG performance. Besides, we compare NL-VSGG\cite{kim2025weaklysupervisedvideoscene}, which uses video captions to construct pseudo scene graphs for model training. NL-VSGG also utilizes an external object detector for object detection, and shows inferior DSGG performance due to the object detection quality issue. We also compare RLIP~\cite{yuan2022rliprelationallanguageimagepretraining} and RLIPv2~\cite{yuan2023rlipv2fastscalingrelational}, which are employed in a zero-shot manner to predict the scene graph by treating each frame as a static image. However, they show inferior performance, which shows the necessity of incorporating temporal context and using relation-aware knowledge for DSGG task.

\subsection{Ablation Study}

In this section, we conduct ablation studies on our method to validate the effectiveness of various designs.

\noindent\textbf{Contribution of knowledge resources.} we validate the effectiveness of detection results from different knowledge resources, which is shown in Tab.~\ref{tab:abla_source}. We can conclude 
(1) adopting ``Internal (w/o IAA)", which denotes class-sensitive attention generated detections, falls behind compared with ``External", which denotes external object detection, indicating that detection-related knowledge obtained from a small number of video scene graph image samples for classification is weaker compared to knowledge pre-trained on large-scale object-centric data. 
(2) ``Internal (w/o IAA) + External", which denotes integrating detection results from different knowledge resources, outperforms solely external detection results by 2.8\% for Average Precision, and 1.48\% for DSGG \textit{With Constraint}, which indicates that knowledge in class-sensitive attention maps provides complementary information for external object detection. 
(3) By adopting our IAA strategy, which utilizes cross-frame cues to augment class-attention maps, the performance is subsequently improved in both ``Internal" and ``Internal+External", showing necessity of temporal cues.

\noindent\textbf{Knowledge Transfer Strategy.} We validate the effectiveness of our knowledge transfer by comparing with other strategies, the results are shown in Tab.~\ref{tab:abla_fusion}.
We can conclude (1) adopting CBM and LRM as knowledge transfer strategies separately yields 1.2\%/2.0\% AP, and consequently gain on DSGG, indicate that object detection quality is improved in terms of bounding box confidence score and location accuracy. (2) By combining CBM and LRM, we harvest the best performance, averaging 2.8\% for AP in object detection and 1.48\%/1.94\% for DSGG under \textit{With Constraint}/\textit{No Constraint}, indicating that box accuracy improvement and confidence score boosting can complement for each other, generating object detection results with higher quality, thus gaining larger performance gain. (3) Incorporating IAA can further boost the detection performance by 8.9\%/10.6\% for AP, indicating it alleviates the blurring and occlusions issue, yielding better detection results, subsequently acquiring better ultimate scene graph generation performance.

\noindent\textbf{Clarification of token types.} We evaluate object detection with different tokens in the class decoder, which are shown in Tab.~\ref{tab:abla_token}, and we apply IAA strategy in this ablation in default. We can conclude (1) Compared with PLA the baseline (Line 1), introducing object tokens  (Line 2) yields performance gain of 9.3\% for AP and 2.6\% for AR, which indicates that object class attention maps indeed highlight regions for each object category, therefore improving the detection results. (2) Further integration of relation tokens (Line 3) harvests 1.8\% performance gain for AP, indicating that by integrating regions containing relation semantics, the class-aware attention maps can highlight regions where objects featuring relations, thus improving performance.

\noindent\textbf{Generalization to Traclet-level WS-DSGG:} To verify the generalization, we also adapt our method to tractlet-level WS-DSGG task, where each video requires one output scene graph with object traclets as nodes. Despite task setting differences, they also use external detectors to generate objects and compose traclets, and face the low-quality detection issues. We choose recent approach UCML~\cite{wu2023UCML} under this setting as baseline and implement our TRKT upon it,
witness improvement on the VidVRD~\cite{shang2017video} dataset, as shown in Tab.~\ref{general}. Though effective in this setting as well, we emphasize our target is frame-level DSGG and such traclet-level setting is out of research scope. 
\begin{table}[!h]
    \centering
    \resizebox{0.32\textwidth}{!}{
    \begin{tabular}{c  ccc}
    \toprule
         Method& \multicolumn{3}{c}{Relation Detection} \\
         \midrule
          & mAP& R@50 & R@100 \\
          \cmidrule{2-4}
         UCML\cite{wu2023UCML}& 17.17\% &  8.48\% & 10.26\% \\
         Ours & \textbf{17.93\%} &  \textbf{9.05\%} & \textbf{11.92\%} \\
         \bottomrule
    \end{tabular}}
    \vspace{-1mm}
    \caption{\textbf{Results on VidVRD~\cite{shang2017video} for traclet-level WS-DSGG.} }
    \label{general}
\end{table}

\input{figures/fig_vis.tex}
\vspace{-4mm}

\subsection{Visualization Results}
\vspace{-2mm}
We provide dynamic scene graph result comparison in Fig.~\ref{vis_dsgg}. Compared with baseline PLA, we yield more complete scene graph with more accurate object localization, which credit to our enhanced object detection quality that integrated with relation-aware temporal knowledge.

\section{Conclusion}

In this work, we recognize the primary challenge for Weakly-Supervised
Dynamic Scene Graph Generation lies in sub-optimal object detection. Therefore, we propose a novel approach TRKT, which firstly mines relation- and motion-aware knowledge tailored for WS-DSGG, then designs Dual-fusion Module to improve the accuracy and confidence score of object bounding boxes, thus enhancing ultimate scene graph’s quality. Our method yields improvement over baseline and achieves sota performance. 
\noindent\textbf{Acknowledgements.} This work was supported by the
grants from the National Natural Science Foundation of
China (62372014, 62525201, 62132001, 62432001), Beijing Nova Program and Beijing Natural Science Foundation (4252040, L247006).


%% file: tables/2_sota_det.tex
\begin{table}[!h]
\Huge
\centering
\resizebox{0.5\textwidth}{!}{
\begin{tabular}{c cc cc}
\toprule
  \multirow{3}{*}{Method}
  & \multicolumn{2}{c}{Average Precision} & \multicolumn{2}{c}{Average Recall}\\
  \cmidrule(lr){2-3} \cmidrule(lr){4-5}
   &maxDets=1 &maxDets=10 &maxDets=1  &maxDets=10 \\
   \midrule
   PLA &11.4 &11.6 &\textbf{32.5} &37.6 \\
   Ours & \textbf{23.0}& \textbf{25.2}&28.8 & \textbf{43.8}\\
\bottomrule
\end{tabular}
}
\vspace{-3mm}
\caption{\textbf{Performance comparison with baseline on Action Genome dataset for object detection.} }
\label{tab:sota_det}
\end{table}

%% file: tables/1_sota_v2.tex
\begin{table}[t]
\centering
\resizebox{0.5\textwidth}{!}{
\begin{tabular}{c ccc ccc}
\toprule
  \multirow{3}{*}{Method}
  & \multicolumn{3}{c}{With Constraint} & \multicolumn{3}{c}{No Constraints}\\
  \cmidrule(lr){2-4} \cmidrule(lr){5-7}

   & R@10 & R@20 & R@50  & R@10 & R@20 & R@50 \\
   \midrule
   RLIP\cite{yuan2022rliprelationallanguageimagepretraining} & 4.72&7.93& 9.16&5.13 & 9.70& 13.80  \\
   RLIPv2\cite{yuan2023rlipv2fastscalingrelational} & 5.06&8.37 &10.05 &5.98 &14.60&21.42\\

   NL-VSGG\cite{kim2025weaklysupervisedvideoscene} & -& 15.75& 20.40& -& 16.11 & 23.21\\

      PLA\cite{chen2023video} & 15.45 & 20.94 &25.79 & 15.87&22.34& 31.69 \\
   Ours & \textbf{17.56}& \textbf{22.33}&\textbf{27.45}& \textbf{18.76}&\textbf{24.49} &\textbf{33.92}\\
   
\bottomrule
\end{tabular}
}
\vspace{-3mm}
\caption{\textbf{Performance comparison with sota methods on Action Genome dataset for WS-DSGG.} }
\label{tab:sota}
\end{table}

%% file: tables/5_abla_source.tex
\begin{table*}[t]
    \centering
    \Huge
    \resizebox{0.85\textwidth}{!}{
    \begin{tabular}{c|c|cc|cc|ccc|ccc}
    \toprule
    \multirow{2}{*}{\makecell[c]{\#}} &
    \multirow{2}{*}{\makecell[c]{Source}} & 
    \multicolumn{2}{c|}{Average Precision} & \multicolumn{2}{c|}{Average Recall} 
  & \multicolumn{3}{c|}{With Constraint} & \multicolumn{3}{c}{No Constraints}\\
    
    & &maxDets=1 &maxDets=10 &maxDets=1  &maxDets=10 
   & R@10 & R@20 & R@50  & R@10 & R@20 & R@50 \\
    \midrule
    1 & Internal (w/o IAA) &2.8 &2.7 &11.2 &11.5  &2.70 &3.02 &3.05 &3.05 &3.99 &4.63 \\
    2 & Internal (w/ IAA) & 5.3& 5.7& 16.9& 18.5 & 5.73& 8.44&13.60 & 5.91& 8.63& 13.71\\
    3 & External &11.4 &11.6 &32.5 &37.6 &14.32 &20.42 &25.43 &14.78 &21.72 &30.87 \\
    4 & Internal(w/o IAA) + External &14.1 &14.6 &\textbf{33.5} &40.4 &16.52 &21.41 &26.68 &17.57 &23.35 &32.27 \\
    5 & Internal(w/ IAA) + External &\textbf{23.0} &\textbf{25.2} &28.8 &\textbf{43.8} &\textbf{17.56}& \textbf{22.33}&\textbf{27.45}& \textbf{18.76}&\textbf{24.49} &\textbf{33.92}\\ 
    
    \bottomrule
    \end{tabular}}
    \vspace{-3mm}
    \caption{\textbf{Ablation study on different knowledge sources.} ``Internal" indicates class-sensitive attention maps generated detection results, ``External" indicates external detection results, ``IAA" denotes adopting Inter-frame augmentation on detection results. }
    
    \label{tab:abla_source}
\end{table*}

%% file: tables/6_abla_fusion.tex
\begin{table*}[t]
    \centering
    \Huge
    \resizebox{0.85\textwidth}{!}{
    \begin{tabular}{c|c|cc|cc|ccc|ccc}
    \toprule
    \multirow{2}{*}{\makecell[c]{\#}} &
    \multirow{2}{*}{\makecell[c]{Knowledge Transfer}} & 
    \multicolumn{2}{c|}{Average Precision} & \multicolumn{2}{c|}{Average Recall} 
  & \multicolumn{3}{c|}{With Constraint} & \multicolumn{3}{c}{No Constraints}\\
    
    & &maxDets=1 &maxDets=10 &maxDets=1  &maxDets=10 
   & R@10 & R@20 & R@50  & R@10 & R@20 & R@50 \\
    \midrule
     1 & - &11.4 &11.6 &32.5 &37.6 &14.32 &20.42 &25.43 &14.78 &21.72 &30.87 \\
    2 & CBM &12.6 &12.3 &33.0 &38.2 &14.87 &21.02 &25.98 &15.41 &22.08 &31.02 \\
    3 & LRM &13.4 &13.7 &33.4 &39.7 &15.50 &21.19 &26.86 &16.01 &22.46 &31.82 \\
    4 & CBM + LRM &14.1 &14.6 &\textbf{33.5} &40.4 &16.52&21.41 &26.68 &17.57 &23.35 &32.27 \\
    5 &CBM+LRM+IAA & \textbf{23.0}& \textbf{25.2}&28.8 & \textbf{43.8}&  \textbf{17.56}& \textbf{22.33}&\textbf{27.45}& \textbf{18.76}&\textbf{24.49} &\textbf{33.92}\\
    \bottomrule
    \end{tabular}}
    \vspace{-3mm}
    \caption{\textbf{Ablation study on knowledge transfer strategy.} ``CBM" and ``LRM" separately represent adopting Confidence Boosting Module and Localization Refinement Module. ``IAA" represents augment attention maps with cross-frame motion.}
    
    \label{tab:abla_fusion}
\end{table*}

%% file: tables/7_abla_token.tex
\begin{table}[t]
\tiny
\centering
\Huge
\resizebox{0.5\textwidth}{!}{
\begin{tabular}{c cc cc}
\toprule
\Huge
  \multirow{2}{*}{Method}
  & \multicolumn{2}{c}{Average Precision} & \multicolumn{2}{c}{Average Recall}\\
  \cmidrule(lr){2-3} \cmidrule(lr){4-5}
   &maxDets=1 &maxDets=10 &maxDets=1  &maxDets=10 \\
   \midrule
   PLA(Baseline) &11.4 &11.6 &\textbf{32.5} &37.6 \\
   PLA + Object  &21.6 &23.0 &31.8 &43.5 \\
   PLA + Object+ Relation  &\textbf{23.0} & \textbf{25.2}& 28.8&\textbf{43.8}\\

\bottomrule
\end{tabular}
}
\caption{\textbf{Ablation study on  different token types.}}
\label{tab:abla_token}
\end{table}

%% file: figures/fig_vis.tex
\begin{figure}[!t]
    \includegraphics[width=0.95\linewidth]{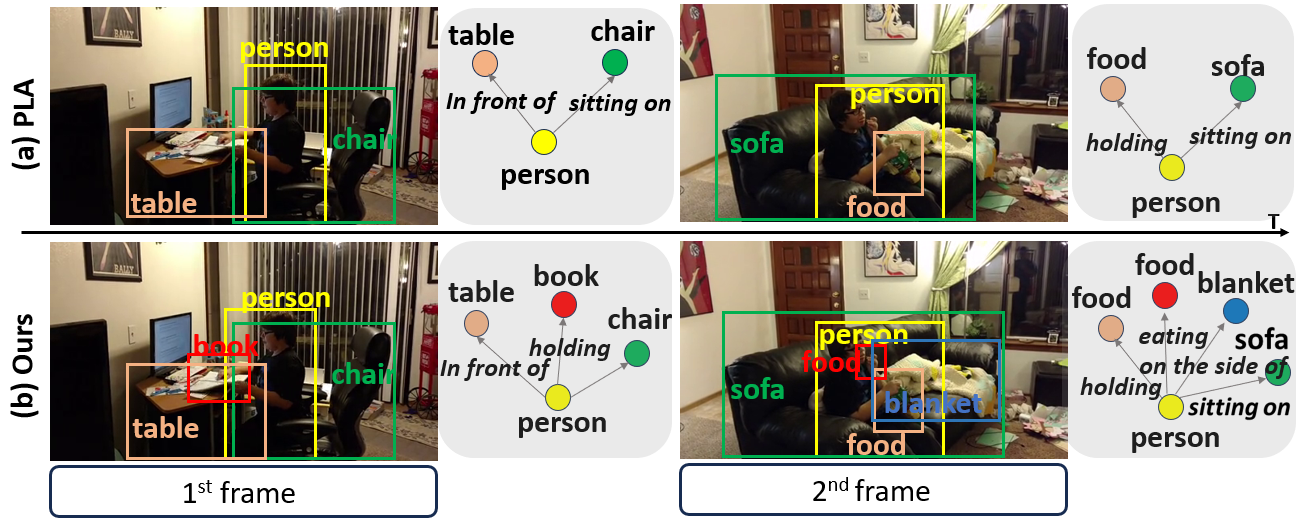}
    \vspace{-2mm}
    \caption{\textbf{Visualization results of dynamic scene graphs.}}
\label{vis_dsgg}
\end{figure}
\vspace{-4mm}